\documentclass[10pt,twocolumn,letterpaper]{article}

\usepackage{cvpr}              

\usepackage{tabularx}    

\definecolor{cvprblue}{rgb}{0.21,0.49,0.74}
\usepackage[pagebackref,breaklinks,colorlinks,allcolors=cvprblue]{hyperref}

\def\paperID{1410} 
\def\confName{CVPR}
\def\confYear{2026}

\title{Loom: Diffusion-Transformer for Interleaved Generation}

\author{
  Mingcheng Ye\textsuperscript{1} \quad
  Jiaming Liu\textsuperscript{2} \quad
  Yiren Song\textsuperscript{3}\textsuperscript{$\dagger$} \\
  \textsuperscript{1}Beijing Institute of Technology \quad
  \textsuperscript{2}Independent Researcher \quad
  \textsuperscript{3}National University of Singapore \\
}

\begin{document}
\twocolumn[{
    \maketitle
    \thispagestyle{empty}
    \begin{@twocolumnfalse}
    \begin{center}
        \vspace{-0.8cm}
        \includegraphics[trim=0.75cm 3.50cm 0.75cm 3.50cm, clip, width=1.0\textwidth, page=2]{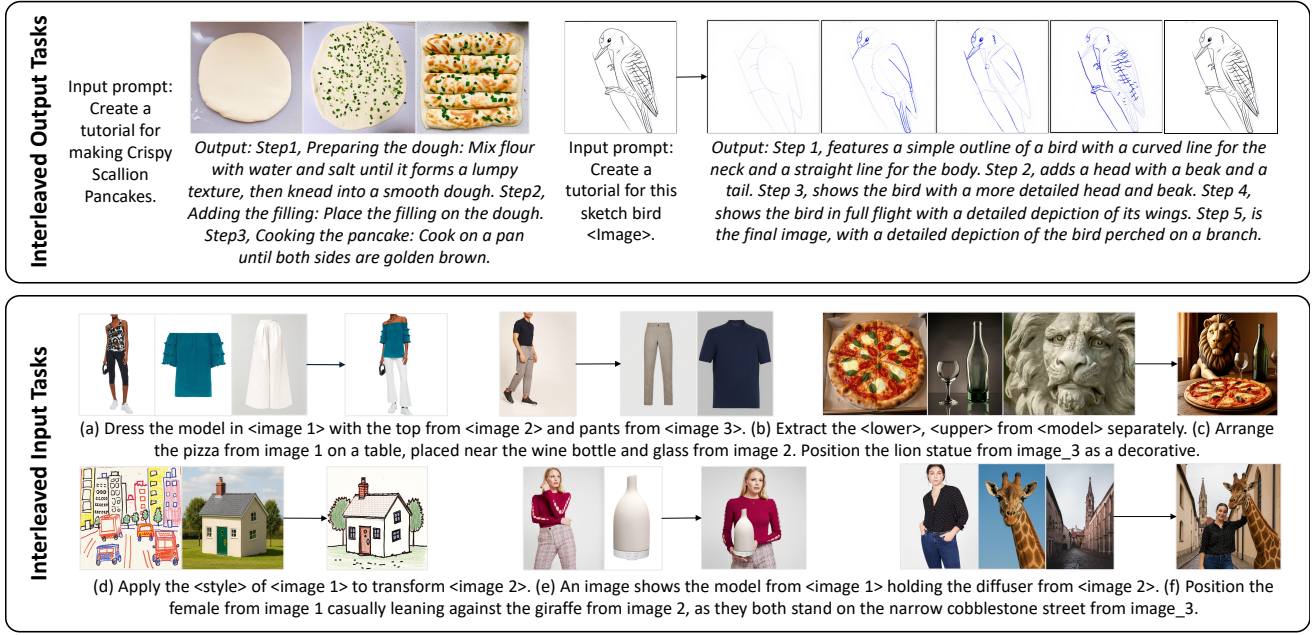}
        \captionof{figure}{Showcases of Loom's interleaved text-image generation. Interleaved input tasks involve composing reference images into one scene or style transfer. Interleaved output tasks produce text–image sequences from a prompt, including cooking tutorials or drawing guide.
        }
        \label{fig:teaser}
    \end{center}
    \vspace{0.2cm}
    \end{@twocolumnfalse}
}]

\begingroup
\renewcommand\thefootnote{}
\footnotetext{$\dagger$ Corresponding author.}
\endgroup

\begin{abstract}

Interleaved text–image generation aims to jointly produce coherent visual frames and aligned textual descriptions within a single sequence, enabling tasks such as style transfer, compositional synthesis, and procedural tutorials. We present Loom, a unified diffusion–transformer framework for interleaved text–image generation. Loom extends the Bagel unified model via full-parameter fine-tuning and an interleaved architecture that alternates textual and visual embeddings for multi-condition reasoning and sequential planning. A language-planning strategy first decomposes a user instruction into stepwise prompts and frame embeddings, which guide temporally consistent synthesis. For each frame, Loom conditions on a small set of sampled prior frames together with the global textual context, rather than concatenating all history, yielding controllable and efficient long-horizon generation. Across style transfer, compositional generation, and tutorial-like procedures, Loom delivers superior compositionality, temporal coherence, and text–image alignment. Experiments demonstrate that Loom substantially outperforms the open-source baseline Anole, achieving an average gain of 2.6 points (on a 5-point scale) across temporal and semantic metrics in text-to-interleaved tasks. We also curate a 50K interleaved tutorial dataset and demonstrate strong improvements over unified and diffusion editing baselines. Code is available at: \href{https://github.com/Plantian/Loom}{https://github.com/Plantian/Loom}

\vspace{-1.00em}
\end{abstract}    
\vspace{-0.50em}
\section{Introduction}
\label{sec:intro}

Open-source unified generative models like Bagel~\cite{deng2025bagel}, Show-O~\cite{xie2025showosingletransformerunify}, Janus-Flow~\cite{ ma2025janusflowharmonizingautoregressionrectified}, BLIP3-o~\cite{chen2025blip3ofamilyfullyopen}, UniWorld~\cite{lin2025uniworldv1highresolutionsemanticencoders} have shown that diverse visual tasks, spanning image editing, stylization, and layout-aware synthesis, can be handled within a single diffusion transformer. Yet most unified systems remain confined to single-turn, single-modality inputs: they either render an image from text or edit one reference in isolation. 

This fundamental limitation, however, extends far beyond simple generation as shown in Fig~\ref{fig:teaser}. A vast and challenging class of real-world scenarios demands reasoning over interleaved, mixed-modality sequences. These N-to-M tasks, which require models to consume and produce multiple, related inputs and outputs, include:
(1) Procedural Generation: Producing step-by-step tutorials where visual frames and textual explanations are interleaved to guide a user, such as in cooking guides or artistic workflows.
(2) Compositional Reasoning: Synthesizing a single, coherent scene from multiple, disparate visual and textual inputs, or the inverse, decomposing a scene into its constituent parts, for applications like virtual try-on.
(3) Multi-Reference Generation: Transforming a content image based on the semantic or stylistic properties of several reference images, such as in complex style transfer. Current open-source frameworks lack a unified mechanism to handle this full spectrum of multi-modal, multi-turn reasoning.

In contrast, proprietary multimodals such as GPT-4o~\cite{openai2024gpt4ocard}, Doubao~\cite{gao2025seedream30technicalreport}, and Gemini~\cite{geminiteam2024gemini15unlockingmultimodal,geminiteam2025geminifamilyhighlycapable} have demonstrated strong proficiency in such interleaved, multi-turn scenarios. These closed-source systems adaptively handle mixed-modality inputs and outputs across conversational contexts, maintaining semantic coherence over extended interactions. However, their proprietary nature restricts research transparency and limits customization for domain-specific applications.

To address this comprehensive gap, we introduce Loom, a unified model that treats text and images as sequentially composable elements within one transformer, enabling interleaved inputs and outputs. Loom is designed to target this full class of N-to-M (N inputs to M outputs) interleaved multi-modal generation tasks. It supports (1) \textbf{Procedural and tutorial generation}, producing step-by-step visual tutorials with aligned text. It handles (2) \textbf{Compositional generation and decomposition}, combining elements for virtual try-on or breaking images into their parts. Finally, it performs (3) \textbf{Style transfer} by conditioning on content images, style references, and text instructions.

To realize this unified approach, Loom treats text and image embeddings as sequentially composable elements within a shared latent space. We introduce a dual set of conditioning mechanisms to manage the complexity of N-to-M tasks. For procedural tasks, a language-planning strategy decomposes global instructions into local steps, which are associated with temporal frame embeddings and sparse historical frame sampling to maintain long-horizon coherence. For compositional and stylistic tasks, control is achieved via learnable entity tokens for structured grounding.

We train Loom by full-parameter fine-tuning on the Bagel backbone, effectively expanding its native capability from single-turn synthesis to complex, multi-turn interleaved generation. A key contribution is our curation of a new 50K interleaved tutorial dataset spanning drawing, cooking, and assembly workflows. Extensive evaluations show that Loom attains high-fidelity interleaved generation with accurate text-image alignment, strong temporal consistency, and robust multi-condition control.
In short, our contributions are as follows:

(1) We propose Loom, a unified diffusion-transformer framework for interleaved text–image generation, supporting style transfer, compositional synthesis, and procedural tutorials within a single model.

(2) We introduce a unified control and conditioning mechanism for N-to-M tasks, including a language-planning strategy and sparse historical frame sampling for temporal coherence, and learnable entity tokens for structured compositional grounding.

(3) We curate a 50K interleaved tutorial dataset and present comprehensive experiments demonstrating Loom’s superior compositionality, temporal coherence, and text–image alignment. By using this dataset, Loom substantially outperforms the open-source baseline Anole, achieving an average gain of 2.6 points (on a 5-point scale) across temporal and semantic metrics in text-to-interleaved tasks.
\section{Related Work}
\label{sec:formatting}

\subsection{Unified Models}
\vspace{-0.20em}
Recent progress toward text--image unified modeling~\cite{wu2024nextgptanytoanymultimodalllm} aims to bridge understanding and generation within a single transformer. Existing frameworks can be broadly categorized by backbone design into diffusion-based models, autoregressive multimodal large language models (MLLM-AR), and MLLM (AR+diffusion) architectures. Diffusion models excel at pixel-level synthesis but often lack bidirectional reasoning for cross-modal understanding. Pure autoregressive MLLMs~\cite{wang2024emu3,chameleonteam2025chameleonmixedmodalearlyfusionfoundation} enable tighter integration between text and image tokens via next-token prediction, yet typically trade off visual fidelity. MLLM (AR+diffusion) frameworks~\cite{xie2025showosingletransformerunify,bai2021pointdsc} inject diffusion decoders into transformer backbones, improving generation while facing potential cross-task interference between modalities.

Our work builds on this unified modeling trend using \textbf{Bagel}~\cite{deng2025bagel}, an MLLM (AR+diffusion) decoder-only transformer with Mixture-of-Experts (MoE) layers for semantic reasoning and visual synthesis. Bagel adopts a Vision Transformer (ViT) based hybrid visual encoder~\cite{tschannen2025siglip2multilingualvisionlanguage} to encode images into visual tokens for multimodal understanding, and a Variational Autoencoder (VAE) with rectified flow~\cite{liu2022rectifiedflowmarginalpreserving,flux2024} for high-fidelity image generation within a shared attention space, providing a strong foundation for the interleaved multimodal tasks studied in this work.

\subsection{Procedural and Tutorial Generation}
\vspace{-0.20em}
Procedural or tutorial generation spans storytelling, creative workflows, and instructional content, but most remains largely text-centric~\cite{ramesh2022hierarchicaltextconditionalimagegeneration, jia2025settleonetexttoimagesetgeneration} without unified bidirectional text–image modeling. Conventional text/image-to-painting pipelines~\cite{chen2024inversepaintingreconstructingpainting,paintsundo,song2025makeanythingharnessingdiffusiontransformers, song2025layertracer} often produce coarse steps lacking aligned textual logic, and recipe datasets like RecipeGen~\cite{zhang2025recipegenbenchmarkrealworldrecipe} yield unimodal outputs. In the realm of unified models, Anole~\cite{chern2024anoleopenautoregressivenative} represents a state-of-the-art open-source baseline but is constrained to text-conditioned inputs. MM-Interleaved~\cite{tian2024mminterleavedinterleavedimagetextgenerative} adds visual tokens but struggles with error accumulation in long horizons. While concurrent approaches like OneFlow~\cite{nguyen2025oneflowconcurrentmixedmodalinterleaved} and Orthus~\cite{kou2025orthusautoregressiveinterleavedimagetext} improve speed, they typically compromise fine-grained stepwise control. 

In contrast, Loom handles both text- and image-conditioned interleaved generation with stable progression, balanced alignment, and precise procedural adherence enabled by our planning-first strategy.

\subsection{Multi-Reference Images Generation}
\vspace{-0.20em}
Interleaved generation necessitates reasoning over multiple visual references simultaneously. One direction is \textbf{style transfer}, where a style reference and a content image yield stylized yet structure-preserving outputs. While diffusion-based approaches~\cite{song2025omniconsistencylearningstyleagnosticconsistency,hertz2024stylealignedimagegeneration} effectively adapt diffusion backbones (e.g., FLUX~\cite{flux2024}) via Low-Rank Adaptation (LoRA)~\cite{hu2021loralowrankadaptationlarge}, they rely heavily on text prompts~\cite{zong2024easyrefomnigeneralizedgroupimage, lei2025stylestudiotextdrivenstyletransfer, wang2024instantstyleplusstyletransfercontentpreserving}, limiting their exploitation of structural guidance from reference images~\cite{wu2025usounifiedstylesubjectdriven,gong2025relationadapter, zhang2024ssr, zhang2024stable, zhang2025stable}. Parallel to this, \textbf{compositional generation and decomposition} require synthesizing coherent scenes from visual context. Although models like Echo-4o~\cite{ye2025echo4oharnessingpowergpt4o} and OmniGen~\cite{xiao2024omnigen,wu2025omnigen2explorationadvancedmultimodal} address fixed configurations, they lack flexibility for decomposition~\cite{wu2025less, wang2025diffdecompose}. Most unified models are restricted to single-round inputs, often struggling with the long-horizon reasoning required for multi-element scenarios.

In contrast, Loom supports long-context multimodal tasks. Our framework scales to accommodate numerous visual inputs and enables diverse applications—ranging from high-fidelity style transfer and accuracy-critical virtual try-on~\cite{chong2025catvtonconcatenationneedvirtual,choi2024improvingdiffusionmodelsauthentic,zhang2024boowvtonboostinginthewildvirtual, guo2025any2anytryon} to commercial model presentation, intricate multi-object arrangement, and reverse component extraction~\cite{zhang2023adding,Zhang2023MagicBrush,cheng2025seedxbuildingstrongmultilingual,wang2025diffdecompose, zhang2024ssr, zhang2025easycontrol}, ensuring semantic consistency across extended sequences.
\begin{figure*}[t]
  \centering
  \includegraphics[trim=0.95cm 5.75cm 0.95cm 2.35cm, clip,width=1.0\textwidth,page=1]{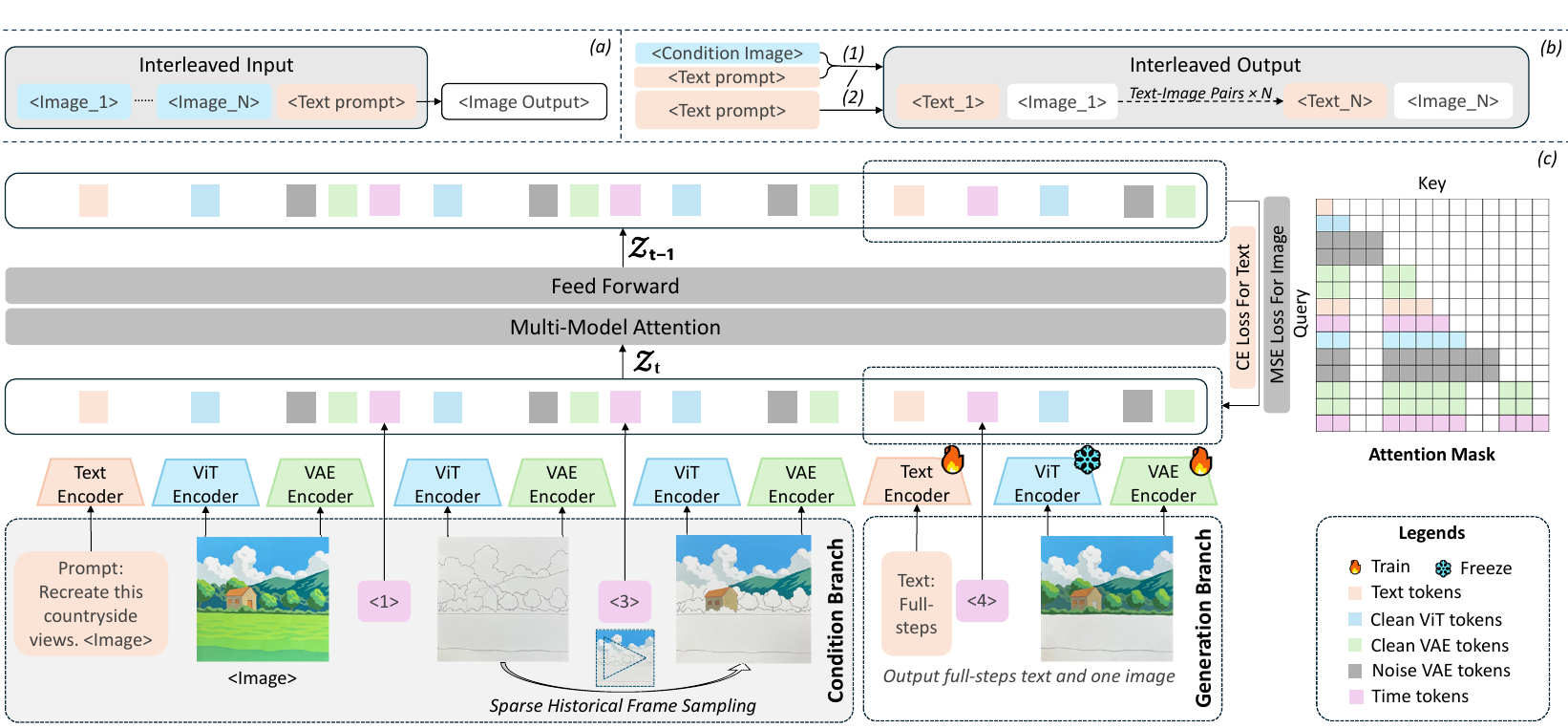}
  \caption{
    \textbf{(a)} An interleaved input paradigm with various conditional images and text prompts, producing a single image output. 
    \textbf{(b)} Interleaved output paradigm, where the model takes either pure text instructions or mixed text-image guidance and generates multi-round, sequential text-image pairs.
    \textbf{(c)} Training and inference architecture for the interleaved output paradigm, focusing on case \textbf{(1)} where the input is a text-image guidance sequence and the output is continuous text-image pairs, exemplified by a step-by-step drawing tutorial. The pipeline contains a \emph{condition branch}, which encodes sparse historical frames via ViT and VAE encoders to provide visual context, and a \emph{generation branch}, which produces both full-step textual descriptions and the next image under the attention mask, ensuring alignment between global textual planning and incremental image rendering.
    }
  \label{fig:architecture}
\end{figure*}

\section{Methods}
\label{sec:method}
The overall architecture of Loom shown in Fig~\ref{fig:architecture}. Our method consists of three components: (1) the Loom model design building on the Bagel backbone with task-specific training configurations; (2) a unified training paradigm with stepwise planning, sparse historical frame sampling, and temporal embeddings for coherent long-horizon multimodal generation; and (3) a curated 50K-sample interleaved dataset covering compositional generation, style transfer, and procedural tutorials. 

\subsection{Overall Architecture}
Loom builds upon the Bagel backbone, which operates autoregressively over interleaved token sequences, where visual content is represented via a pre-trained VAE encoder and decoded through the rectified flow method. Following Bagel, Loom unifies text prediction and image generation within a single transformer framework.

\vspace{-1.00em}
\paragraph{Multi-Modal Attention.}
Loom inherited Bagel's multi-modal attention (MMA) mechanism to provide conditional information for both text and image token generation. The MMA formulation is defined as:

\vspace{-1.50em}
\begin{equation}
\mathrm{MMA}([c_T; c_Z; c_I]) = \mathrm{softmax}\left(\frac{QK^\top}{\sqrt{d}}\right) V
\end{equation}
\vspace{-0.90em}

where $[c_T; c_Z; c_I]$ denotes the concatenation of three types of tokens: (1) \textbf{$c_T$}: text tokens, encoding semantic or instructional information. (2) \textbf{$c_Z$}: noised latent tokens, representing partially corrupted visual representations to be denoised. (3) \textbf{$c_I$}: image condition tokens, providing visual context or reference frames.

\vspace{-1.00em}
\paragraph{Interleaved Training Objects.}
We adopt the standard Bagel training objectives for interleaved synthesis, combining language modeling for text~\cite{bai2025qwen25vltechnicalreport} tokens and rectified flow matching~\cite{lipman2023flowmatchinggenerativemodeling,liu2022flowstraightfastlearning} for image tokens. The overall training loss is:

\vspace{-0.60em}
\begin{equation}
    \mathcal{L}_{\text{total}} = \lambda_{\text{CE}} \cdot \mathcal{L}_{\text{CE}}^{\text{text}} + \mathcal{L}_{\text{MSE}}^{\text{image}}
\end{equation}
\vspace{-0.60em}

Here, $\mathcal{L}_{\text{CE}}^{\text{text}}$ denotes the cross-entropy loss for next-token text prediction, and $\mathcal{L}_{\text{MSE}}^{\text{image}}$ denotes the mean squared error loss used for image denoising in latent space. The coefficient $\lambda_{\text{CE}}$ controls the relative weight between textual and visual objectives. This unified objective allows Loom to jointly optimize heterogeneous interleaved tasks within a single autoregressive framework.

\subsection{Interleaved Text Image Generation Tasks}
Our training paradigm unifies heterogeneous interleaved tasks under a single autoregressive framework through three key algorithmic designs.

\vspace{-1.00em}
\paragraph{Planning-First Strategy.}
\label{para:stepwiseplaning}
Standard interleaved models typically alternate between text and image generation steps ($T_1 \to I_1 \to T_2 \dots$), where errors in early visual frames ($I_1$) propagate to subsequent text predictions ($T_2$), causing semantic drift~\cite{liao2025mogaoomnifoundationmodel} in long-horizon tasks. To mitigate this, we propose a Planning-First Strategy that decouples high-level semantic reasoning from pixel-level synthesis.

Inspired by Chain-of-Thought reasoning~\cite{qin2025unicotunifiedchainofthoughtreasoning,li2025zebracotdatasetinterleavedvision}, Loom adopts a "plan-then-render" workflow. Given a multimodal instruction, the model first generates the entire formatted textual plan $\mathcal{P} = \{S_1, S_2, \dots, S_N\}$ in a single autoregressive pass, decomposing the task into coherent sub-steps (e.g., \texttt{Step~1: prepare ingredients... Step~N: final plating}) without visual interference. Once the plan is fixed, Loom enters a deterministic rendering loop: for each step $t$, the image $I_t$ is generated conditioned on (1) the clean textual description $S_t$, (2) the global plan context $\mathcal{P}$, and (3) selected historical frames. For each frame $I_t$ in an $N$-step sequence, we construct a training instance that predicts \emph{both} the complete plan $\mathcal{P}$ and the current image $I_t$ sequentially: the model first generates $\mathcal{P}$ autoregressively, then renders $I_t$ conditioned on the corresponding step description extracted from $\mathcal{P}$. This multi-round alignment—where the same plan is trained with different visual states—ensures global coherence across all frames. Crucially, as shown in Fig.~\ref{fig:architecture}(c), noised VAE tokens (gray blocks) do not participate in attention; only clean historical frames (green blocks) condition the generation. The noise isolation design prevents the model from learning spurious correlations with intermediate diffusion states, effectively preventing noisy visual states from contaminating the logical progression of the text.

\vspace{-0.80em}
\paragraph{Temporal Embeddings.}
To further stabilize long-horizon interleaved generation, we incorporate a \emph{learnable time embedding} that explicitly encodes the relative position of each visual frame within the planned generation sequence. For a frame at step $t$, we add a dedicated vector $\mathbf{e}_t$ to its visual tokens before multimodal attention:

\vspace{-0.60em}
\begin{equation}
    \tilde{\mathbf{x}}_t = [c_T; c_{I,Z}^{(t-1)}; \mathbf{v}_t] + \mathbf{e}_t,
\end{equation}
\vspace{-0.60em}

where $\mathbf{v}_t$ denotes the visual tokens of the $t$‑th frame (obtained via VAE/ViT encoding), $c_T$ represents the textual tokens providing semantic instructions, and $c_{I,Z}^{(t-1)}$ are image‑condition tokens carried over from the previous step $t{-}1$. $\mathbf{e}_t$ is a learnable temporal embedding for step $t$, added before multimodal attention to inform the model of the current frame’s position in the generation sequence and to strengthen sequential coherence.

\vspace{-1.00em}
\paragraph{Sparse Historical Frame Sampling.}
\label{para:sampling}
Conditioning on a fixed sliding window of recent frames fails to capture global context in long sequences. To balance computational efficiency with global temporal awareness, we introduce a deterministic Uniform Sparse Sampling mechanism. 

Given a sequence of $t$ prior frames and a maximum reference budget $K_{\max}$, we select a subset of frames at indices:

\vspace{-0.30em}
\begin{equation}
k_i = \left\lfloor \frac{i \cdot t}{K_{\max}+1} \right\rfloor, \quad i \in \{1, \dots, K_{\max}\}
\end{equation}

Each selected frame is encoded via dual pathways—a ViT for high-level semantic features and a VAE for fine-grained visual details—before being injected into the Multi-Modal Attention block. 

This uniform sampling strategy offers three key advantages: (1) Computational efficiency—maintaining $O(K_{\max})$ complexity for all frames; (2) Reduced error propagation—skipping intermediate frames avoids accumulating visual artifacts; (3) Long-range structure preservation—sampling at uniform intervals captures both initial state and recent changes simultaneously. For each selected frame, visual features are augmented with a learnable temporal embedding $\mathbf{e}_{t}$ representing its absolute step index, allowing the model to integrate long-range scene structure and short-term visual detail with negligible overhead.

\subsection{Entity-Anchored Control via Localized CFG}
\label{sec:entity}

Standard text prompts lack explicit handles for individual objects, making fine-grained control difficult in compositional tasks. We introduce learnable entity tokens (e.g., $\langle \text{model} \rangle$, $\langle \text{garment} \rangle$) that serve as semantic anchors immediately followed by detailed descriptions of the corresponding objects. For example: \texttt{"a $\langle$model$\rangle$ [in casual pose] wearing a $\langle$garment$\rangle$ [red floral dress]"}. This structure allows the model to bind entity tokens to specific visual concepts during training, while their subsequent descriptions provide fine-grained attributes, enabling both compositional generation (fusing entities) and decomposition (isolating entities) within a unified framework. Clean object isolation in our dataset enables precise entity-visual associations without complex foreground-background separation. By leveraging Bagel's classifier-free guidance (CFG)~\cite{ho2022classifierfreediffusionguidance}, we train the model to learn both conditional $p(\mathbf{x} \mid \mathbf{c})$ and unconditional $p(\mathbf{x})$ distributions by randomly dropping conditions during training. At inference, CFG modulates the generation via:

\vspace{-1.80em}
\begin{equation}
\begin{aligned}
\nabla_{\mathbf{x}} \log p(\mathbf{x} \mid \mathbf{c}) = 
\gamma \Big(&
\nabla_{\mathbf{x}} \log p(\mathbf{x} \mid \mathbf{c}) - \nabla_{\mathbf{x}} \log p(\mathbf{x})\Big) \\ + 
\nabla_{\mathbf{x}} \log p(\mathbf{x})
\end{aligned}
\end{equation}
\vspace{-1.50em}

where $\gamma$ is Bagel's $cfg\_text\_scale$. The learned entity tokens concentrate attention on their associated visual regions through cross-attention mechanisms trained on our compositional dataset, where clean object isolation provides implicit spatial supervision. This enables entity-aware control without explicit spatial masks or auxiliary segmentation models.

\subsection{Interleaved Dataset Construction}
\begin{figure}[t]
\begin{minipage}{0.50\textwidth}
    \centering
    \includegraphics[trim=1.15cm 11.10cm 14.50cm 0.15cm, clip,width=\textwidth,page=13]{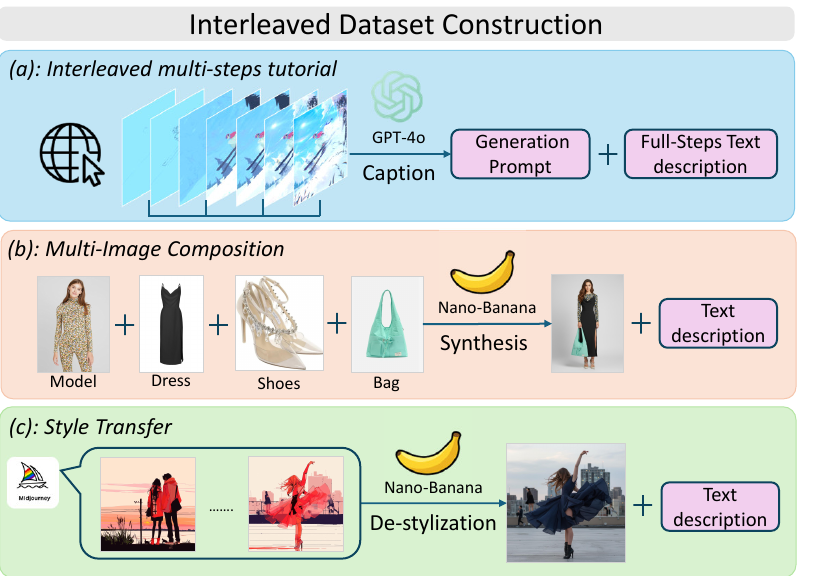}
    \captionsetup{width=1.0\textwidth}
    \caption{
      Interleaved dataset construction: (a) Blogs and videos are collected, and 4–6 frames are uniformly sampled, manually verified, and captioned by GPT‑4o with generation prompts and stepwise captions. (b) Multi‑image composition combines models, objects, and scenes via Nano‑Banana~\cite{Nano-banana} with textual descriptions. (c) Style transfer uses Promptsref~\cite{promptsref} website images; Nano‑Banana performs de‑stylization to obtain realistic images and corresponding prompts.
    }
    \label{fig:interleaveddatasetconstrcution}
\end{minipage}\hfill
\end{figure}

Creating a robust interleaved generator requires data that excels in both temporal logic and text-image alignment. As shown in Fig~\ref{fig:interleaveddatasetconstrcution}, we construct a high-fidelity 50K-sample dataset through a rigorous multi-stage pipeline spanning three distinct task categories.

\vspace{-1.00em}
\paragraph{Compositional Generation and Decomposition.}
We curate a logically complex subset from Echo-4o\cite{ye2025echo4oharnessingpowergpt4o} and our internal imagery (11k, generated by Nano-banana~\cite{Nano-banana}), focusing on multi-object interactions. To ensure robust spatial reasoning, we systematically perform entity-level permutation and combination across categories (models, garments, products). Unlike standard datasets, we generate paired bi-directional instructions: forward commands for integrating entities into unified layouts, and reverse commands for decomposing scenes into constituent parts. All samples undergo strict overlap checks to remove redundancy, yielding a diverse corpus that mirrors real-world virtual try-on and layout synthesis scenarios.

\vspace{-1.00em}
\paragraph{Style Transfer Pairs.} 
We construct a paired stylization dataset using 12k high-aesthetic references from Promptsref~\cite{promptsref}. To enforce content-structure disentanglement, we employ a "De-stylization" pipeline using Nano-Banana to generate photorealistic content counterparts for each stylized reference. This results in perfectly aligned (Content, Style, Output) triplets, enabling the model to learn precise style injection while preserving structural fidelity under supervised conditions.

\vspace{-1.00em}
\paragraph{Sequential Procedural Tutorials.}
We focus on long-horizon logic in cooking and creative arts. For cooking, we extract a high-alignment subset (14k) from RecipeGen\cite{zhang2025recipegenbenchmarkrealworldrecipe}, filtering for sequences with $\geq{3}$ steps where visual state changes strictly follow textual instructions. For digital painting, we build a specialized 8k dataset from iPad-based screen recordings, manually filtered to reflect human cognitive progression (sketch $\to$ color $\to$ detail). Additionally, we reprocess a 5k subset from MakeAnything\cite{song2025makeanythingharnessingdiffusiontransformers}, using GPT-4o to refine stepwise captions for causal consistency. This tiered filtration ensures that every frame transition in our training data serves as a valid logical progression for the planner.

\label{sec:main_experiment}
\section{Experiment}
\begin{figure*}[t]
  \centering
  \includegraphics[trim=0.00cm 4.85cm 0.55cm 0.50cm, clip, width=1.0\textwidth, page=1]{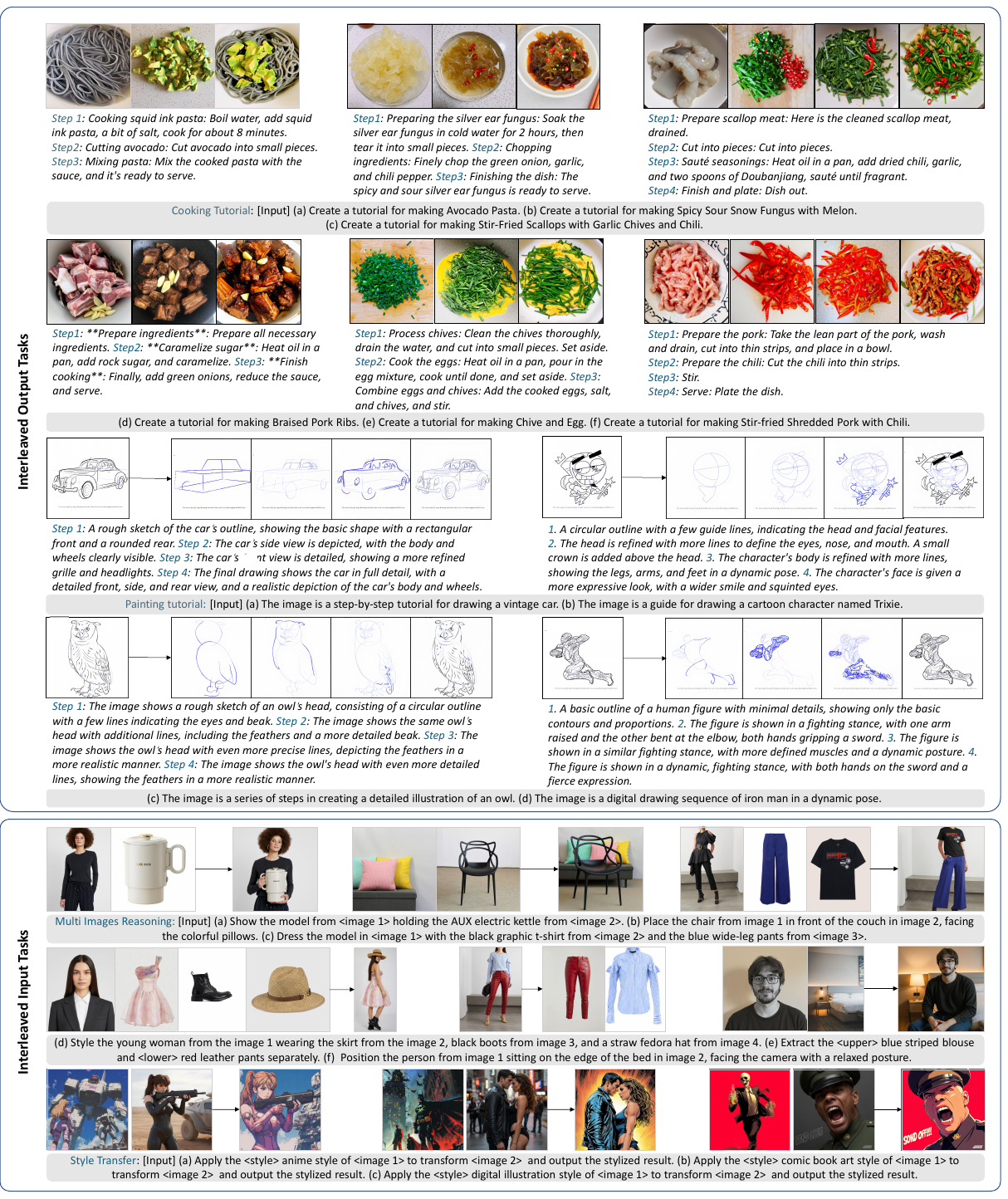}
  \caption{More generation results of Loom in interleaved tasks. 
    The top rows show interleaved output tasks, including text-to-interleaved cooking tutorials and image-to-interleaved painting tutorials. 
    The bottom rows depict interleaved input tasks, covering (1) procedural generation, (2) compositional generation and decomposition, and (3) style transfer from a given reference image.}
  \label{fig:results}
\end{figure*}
\subsection{Set up}

\paragraph{Implementation details.}
We employ the supervised fine-tuning (SFT) method in Bagel post-training tasks; we set the learning rate $2.5\times 10^{-5}$ with a constant scheduler. Optimization uses AdamW with a CE to MSE loss weight ratio of $0.25:1$. The total training steps are set as follows: 16{,}000 steps for the compositional generation (long-text, multi-image reasoning) task, 10{,}000 steps for style transfer, 20{,}000 steps for recipe-oriented text–image interactive generation, and 5{,}000 steps for interleaved painting tutorials. The batch size is fixed at 4. We used 4\,NVIDIA H200 GPUs for full-parameter fine-tuning of Bagel in training Loom.

\vspace{-1.10em}
\paragraph{Benchmarks.}
For multi-image reasoning, we adopt the OmniContext benchmark, which evaluates the ability to integrate multiple visual references by measuring scene completeness, object fidelity, and semantic consistency. For text-to-interleaved generation, we evaluate on 50 cooking recipes from RecipeGen test dataset. For image-to-interleaved generation, we randomly select 20 painting tutorials from MakeAnything. 

\vspace{-1.50em}
\paragraph{Metrics.}
We evaluate interleaved generation in two settings: image-conditioned and text-only. Performance is measured across three dimensions: \emph{visual continuity} (temporal coherence between frames), \emph{textual continuity} (logical flow of descriptions), and \emph{cross-modal alignment} (consistency between images and instructions). For text-to-interleaved generation, we also report the CLIP~\cite{radford2021learningtransferablevisualmodels} (ViT-L/14) score to quantify prompt–image alignment. Human evaluation is conducted by 20 annotators; GPT-4o evaluation uses identical rating scales. Detailed evaluation protocols, reliability and correlation analysis are provided in Appendix~\ref{sec:EMR}.

\vspace{-1.50em}
\paragraph{Baseline methods.}
For the text–image interleaved generation task, we compare our approach with Anole. We also adapt unified models such as Janus-Pro and Bagel by constructing a multi-turn dialogue framework to support interleaved outputs, as these models are originally designed for single-turn generation. To enable fair comparison, we provide these adapted baselines with step-by-step textual prompts (generated by GPT-4o, identical to Loom's 
planning output) at each generation step. In addition, we include a proprietary large multimodal model, Doubao, as a reference baseline due to its support for high-quality interactive text–image generation.

\subsection{Quantitative Comparison}
\label{part:comparison}
\begin{table}[!htbp]
\centering
\scriptsize 
\caption{Text-to-interleaved results using GPT-4o(G) and Human(H) scoring. Coh. = Temporal Coherence; Ins. = Instruction Following; Con. = Narrative Consistency; CLIP = CLIP Score; G = GPT-4o score; H = Human score.}
\vspace{-1.00em}
\label{tab:interleavetextgen}
\renewcommand{\arraystretch}{1.0}
\begin{tabularx}{0.475\textwidth}{
    >{\centering\arraybackslash}p{1.4cm}  
    | >{\centering\arraybackslash}p{1.4cm} 
    | >{\centering\arraybackslash}p{1.4cm} 
    | >{\centering\arraybackslash}p{1.4cm}
    | >{\centering\arraybackslash}p{0.6cm}
}
\toprule
\textbf{Model} & \textbf{Coh.(G \textbar\space H)$\uparrow$} & \textbf{Ins.(G \textbar\space H)$\uparrow$} & \textbf{Con.(G \textbar\space H)$\uparrow$} & \textbf{CLIP$\uparrow$}\\
\midrule
Doubao~\cite{gao2025seedream30technicalreport} & \textbf{4.35 \textbar\space 4.10} & \textbf{4.25 \textbar\space 4.05} & \textbf{4.95 \textbar\space 4.65} & 0.250 \\
\midrule
Bagel~\cite{deng2025bagel} & 1.40 \textbar\space 1.25 & 1.55 \textbar\space 1.05 & --    & 0.217 \\
Janus-Pro~\cite{chen2025januspro} & 1.05 \textbar\space 1.10 & 1.10 \textbar\space 1.00 & --    & 0.105 \\
Anole~\cite{chern2024anoleopenautoregressivenative} & 1.55 \textbar\space 1.05 & 1.35 \textbar\space 1.05 & 1.95 \textbar\space 1.35 & 0.219 \\
\textbf{Loom (Ours)} & \textbf{4.25 \textbar\space 4.15 } & \textbf{3.75 \textbar\space 3.35 } & \textbf{4.70 \textbar\space 4.30} & \textbf{0.269} \\
\bottomrule
\end{tabularx}
\end{table}

\begin{table}[!htbp]
\centering
\scriptsize 
\caption{Image-to-interleaved results using GPT-4o and human scoring. Coh. = Temporal Coherence; Ref. = Reference Faithfulness; Ali. = Semantic Alignment; G = GPT-4o score; H = Human score.}
\vspace{-1.00em}
\label{tab:interleavetextpic}
\renewcommand{\arraystretch}{1.0}
\begin{tabularx}{0.475\textwidth}{
    >{\centering\arraybackslash}p{1.9cm}  
    | >{\centering\arraybackslash}p{1.5cm}
    | >{\centering\arraybackslash}p{1.5cm} 
    | >{\centering\arraybackslash}p{1.5cm} 
}
\toprule
\textbf{Model} & \textbf{Coh. (G \textbar\space H)$\uparrow$} & \textbf{Ref. (G \textbar\space H)$\uparrow$} & \textbf{Ali. (G \textbar\space H)$\uparrow$} \\
\midrule
Doubao & 2.05 \textbar\space 2.15 & 2.65 \textbar\space 2.65 & \textbf{3.55 \textbar\space 3.85} \\
\midrule[0.2pt]
Bagel & 1.25 \textbar\space 1.00 & 1.15 \textbar\space 1.55 & -- \\
\textbf{Loom (Ours)} & \textbf{3.15 \textbar\space 3.65} & \textbf{3.85 \textbar\space 4.15} & \textbf{3.15 \textbar\space 2.95} \\
\bottomrule
\end{tabularx}
\end{table}

\paragraph{Task~1: Interleaved generation.}
We evaluate two sub-settings: text-to-interleaved (Table~\ref{tab:interleavetextgen}) and image-to-interleaved (Table~\ref{tab:interleavetextpic}). Compared to the baseline Bagel, Loom achieves an average relative improvement of over 50\% in text-to-interleaved generation and around 44\% in image-to-interleaved generation, while outperforming the strongest open-source interleaved model Anole by over 51\% in the text setting. \emph{(Con. and Ali. are not reported for baseline model Bagel and Janus-Pro that do not produce paired text-image guidance in the given setting.)}

In the text-to-interleaved setting, extending Bagel with our multi-turn framework enables continuous stepwise generation from scratch, where the native model cannot. Across metrics, Loom far surpasses Anole and all single-turn unified baselines. In the image-to-interleaved setting, Loom achieves the highest average scores across all metrics (3.48 vs. Doubao's 2.82), excelling particularly in temporal coherence (+62\%) and reference faithfulness (+51\%). While Doubao shows better semantic alignment in most cases, Loom's superior visual consistency and detail preservation make it more suitable for long-horizon procedural tasks.

\begin{figure}[t]
\begin{minipage}{0.48\textwidth}
    \centering
    \includegraphics[trim=0.30cm 13.00cm 10.15cm 0.50cm, clip,width=1.00\textwidth,page=2]{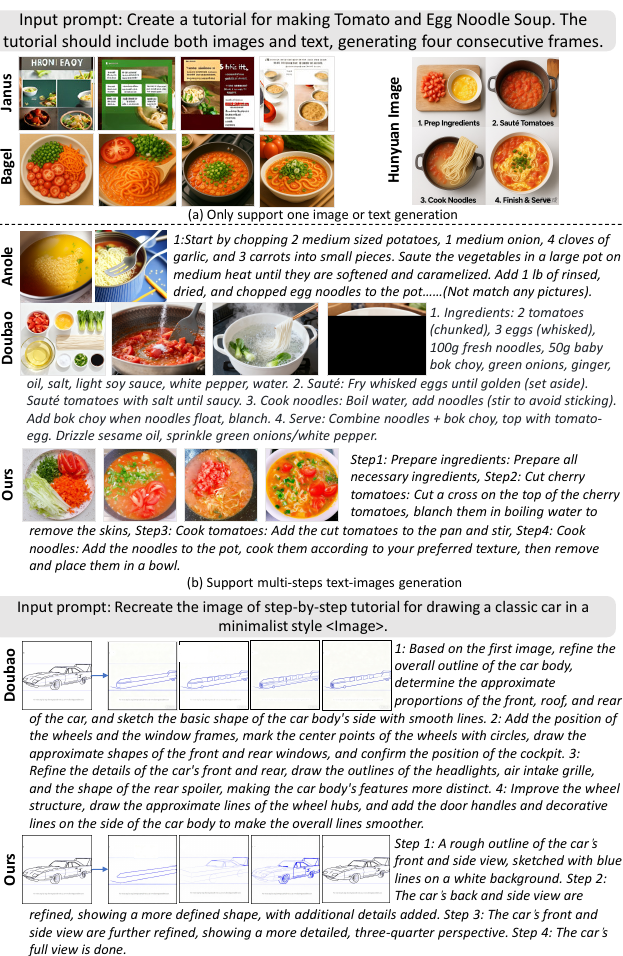}
    \caption{Comparison results. 
    (a) Unified models such as Bagel and Janus-Pro only support single-round input–output generation. 
    (b) Interleaved models, including Anole and Doubao support multi-step text–image generation; we also compared closed-source Doubao with our Loom in both text- and image-to-interleaved tasks.
    }
    \label{fig:comparisonresults}
\end{minipage}\hfill
\end{figure}

Qualitative comparisons in Figure~\ref{fig:comparisonresults} show that open-source single-turn unified models quickly lose logical progression in later frames, while Hunyuan Image Edit fail to preserve details. Anole struggles with text–image alignment; Doubao produces visually appealing and well-aligned results but still exhibits small logical breaks, whereas Loom maintains superior alignment, coherence, and near-perfect adherence to reference images in painting tasks.

\vspace{-0.50em}
\begin{table}[!htbp]
\centering
\scriptsize 
\caption{Multi-image reasoning performance on the OmniContext benchmark. Scores are categorized by scene type (MULTIPLE and SCENE) and the composition of "Character" (Char.), "Object" (Obj.), and both (Char.+Obj.). Higher average scores (Avg) indicate better performance. The benchmark's SINGLE category is excluded, as it is not relevant to our multi-image task.}
\label{tab:omnibenchmark}
\renewcommand{\arraystretch}{1.0}
\setlength{\tabcolsep}{3.0pt}
\begin{tabular}{c|ccc|ccc|c}
\toprule
\textbf{Model} & \multicolumn{3}{c|}{\textbf{MULTIPLE}} & \multicolumn{3}{c|}{\textbf{SCENE}} & \textbf{Avg$\uparrow$} \\
\cline{2-7}
& Char. & Object & Char.+Obj. & Char. & Object & Char.+Obj. & \\
\midrule
Gemini-2.0-flash & 2.91 & 2.16 & 3.80 & 3.02 & 3.89 & 2.92 & 3.12 \\
GPT-4o~\cite{openai2024gpt4ocard} & \textbf{9.07} & \textbf{8.95} & \textbf{8.54} & \textbf{8.90} & \textbf{8.44} & \textbf{8.60} & \textbf{8.75} \\
\midrule[0.2pt]
UNO~\cite{wu2025less}  & 2.54 & 6.51 & 4.39 & 2.06 & 4.33 & 4.37 & 4.03 \\
Bagel~\cite{deng2025bagel} & 5.17 & 6.64 & 6.24 & 4.07 & 5.71 & 5.47 & 5.55 \\
OmniGen~\cite{xiao2024omnigen} & 5.65 & 5.44 & 4.68 & 3.59 & 4.32 & 5.12 & 4.8 \\
OmniGen2~\cite{wu2025omnigen2explorationadvancedmultimodal} & 7.11 & 7.13 & 7.45 & 6.38 & 6.71 & 7.04 & 6.97 \\
Echo-4o~\cite{ye2025echo4oharnessingpowergpt4o} & 8.07 & 7.50 & 8.29 & 8.62 & 8.00 & 8.08 & 8.09 \\
\textbf{Loom(Ours)} & \textbf{8.09} & \textbf{7.62} & \textbf{8.25} & \textbf{8.67} & \textbf{7.95} & \textbf{8.23} & \textbf{8.13} \\
\bottomrule
\end{tabular}
\end{table}

\paragraph{Task~2: Compositional Generation (multi-image reasoning).}
For this task, performance is directly reported using the OmniContext benchmark in Table~\ref{tab:omnibenchmark}.

\subsection{Qualitative Evaluation}
More qualitative generation results are shown in Figure~\ref{fig:results}. The outputs demonstrate Loom is the current open-source SOTA in interleaved generation.

\subsection{Ablation}
\begin{figure}[t]
\begin{minipage}{0.48\textwidth}
    \centering
    \includegraphics[trim=0.00cm 16.00cm 7.50cm 0.00cm, clip,width=1.0\textwidth,page=3]{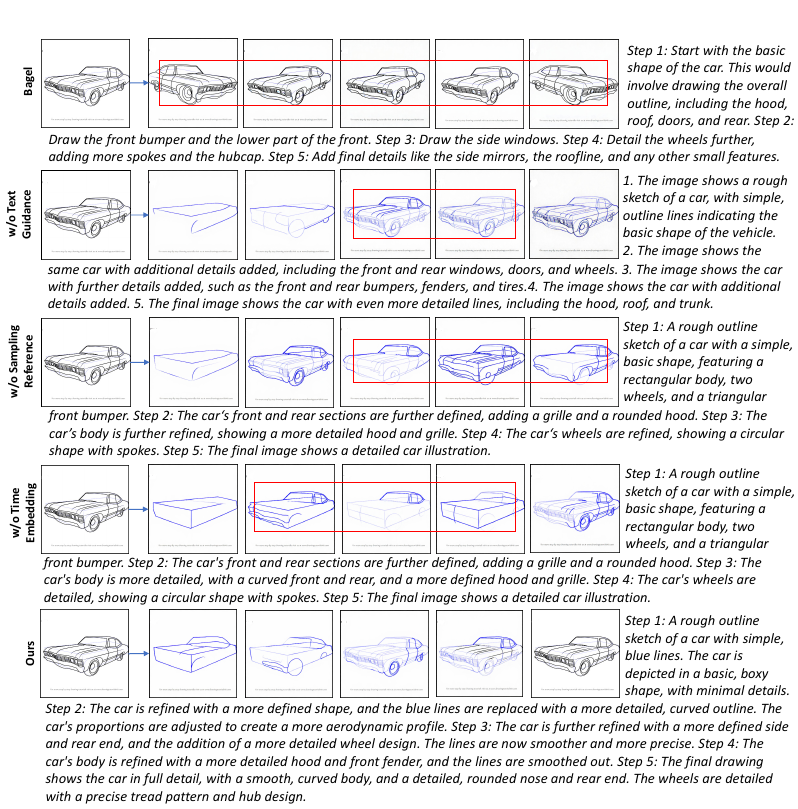}
    \caption{Ablation study results. From top to bottom: (1) baseline (Bagel), (2–4) removing each of the three modules: stepwise textual guidance, reference sampling, and time embedding and (5) our full model Loom. Red rectangles highlight regions with obvious errors.}
    \label{fig:ablationstudyforinterleavedresults}
\end{minipage}\hfill
\end{figure}

We perform ablation studies exclusively on the image-to-interleaved generation setting (painting tutorials). More ablation results are in Figure~\ref{fig:ablationstudyforinterleavedresults}. The ablation results in Table~\ref{tab:abalationfort2inter} reveal contributions from all components. Removing any single module leads to a clear drop across at least two metrics, confirming their complementary roles: textual guidance provides fine-grained control over semantic alignment, reference sampling strengthens temporal coherence via richer visual context, and time embedding is critical for maintaining consistent progression across frames. Compared to the baseline model Bagel, the full system improves temporal coherence by +38\%, reference faithfulness by +54\%, underscoring the necessity of all components.

\begin{table}[!htbp]
\centering
\scriptsize 
\caption{Ablation Study for Painting Tutorials (Image-to-Interleave) with GPT-4o Evaluation. Coherence. = Temporal Coherence; Reference. = Reference Faithfulness; Alignment. = Semantic Alignment.}
\label{tab:abalationfort2inter}
\renewcommand{\arraystretch}{0.9} 
\setlength{\tabcolsep}{1.0pt}        
\begin{tabularx}{0.475\textwidth}{
    >{\centering\arraybackslash}p{3.0cm}  
    | >{\centering\arraybackslash}p{1.6cm}
    | >{\centering\arraybackslash}p{1.6cm} 
    | >{\centering\arraybackslash}p{1.6cm} 
}
\toprule
\textbf{Method} & \textbf{Coherence.$\uparrow$} & \textbf{Reference.$\uparrow$} & \textbf{Alignment.$\uparrow$} \\
\midrule
Baseline & 1.25 & 1.15 & -- \\
w/o Time Embedding & 2.55 & 2.35 & 2.85 \\
w/o Stepwise Prompt & 2.15 & 2.95 & 2.35 \\
w/o Reference Sampling & 1.45 & 1.25 & 2.05 \\
\textbf{Full} & \textbf{3.15} & \textbf{3.85} & \textbf{3.15} \\
\bottomrule
\end{tabularx}
\end{table}

\section{Conclusion}
\label{sec:conclusion}

We present \textbf{Loom}, a unified diffusion–transformer framework for interleaved text–-image generation across style transfer, compositional synthesis, and procedural tutorials. Built by extending the Bagel pre-trained unified model from single-turn inputs to handle complex N-to-M interleaved sequences, Loom supports interleaved procedural sequencing, structured compositional manipulation, and reference-guided appearance transfer. Experiments on a 50K interleaved dataset demonstrate that Loom delivers superior compositionality, temporal coherence, and text-–image alignment, significantly outperforming open-source baselines and setting a new benchmark for coherent interleaved multimodal generation.
{
    \small
    \bibliographystyle{ieeenat_fullname}
    \bibliography{main}
}


\clearpage
\setcounter{page}{1}
\maketitlesupplementary
\appendix
\renewcommand{\thesection}{\Alph{section}}
\setcounter{section}{0}
\section{Implementation Details}

\subsection{Method Details}
\paragraph{Training Scaffolding for Planning.}
To enable the Planning-First strategy described in the main text, we construct specific context scaffolds during training. Given a raw prompt (optionally with a reference image), we prepend a lightweight skeleton instruction to guide the model. Conditioned on this scaffold, the model learns to output the full plan in a standardized format. The maximum number of steps is truncated at $N_{max}=6$ during training to fit memory constraints, though the model generalizes to variable lengths at inference. This consistent formatting ensures that the "Plan" phase is robust and distinct from the "Render" phase.

\paragraph{Sampling Rotation Offset.}
While the inference phase uses the deterministic sparse sampling formula introduced in Sec.~\ref{para:sampling}, during training, we introduce a randomization factor to improve robustness. We apply a small rotation offset $o$ to the sampling indices:
\begin{equation}
    \text{Indices} = \{ \lfloor i \cdot T / (K_{max}+1) \rfloor + o \mid i=1,\dots,K_{max} \}
\end{equation}
where $0 \leq o < \lfloor T/(K_{max}+1) \rfloor$. This offset shifts the sampled frames slightly across different training epochs.
\textbf{Example:} For a sequence length $T=5$ with budget $K_{max}=4$:
\begin{itemize}
    \item $o=0 \rightarrow \{1,2,3,4\}$
    \item $o=1 \rightarrow \{2,3,4,5\}$
\end{itemize}
This mechanism ensures that over the course of training, every frame position in the history is observed by the model at least once, preventing overfitting to specific step indices. At inference, we default to full-context mode if $T$ permits, or use the standard deterministic sampling ($o=0$) for longer sequences to maximize stability.

\paragraph{Entity Tokens Details.}
Our dataset is curated so each frame contains a single, clean foreground subject (person or object). We introduce a small vocabulary of textual entity tokens (e.g., \texttt{<model>}, \texttt{<upper>}, \texttt{<bag>}) that canonically name the subject. These tokens recur across diverse scenes, poses, garments, and styles, enabling the model to learn stable, compositional grounding to real-world entities.

During training, the source images are intentionally constructed to depict a single salient object or person per frame. To leverage this natural compositional unit, Loom augments textual conditioning with structured entity tokens that explicitly represent the main subject in each image.

At inference, entity tokens serve as handles for localized control. Inserting a token into the plan (e.g., “replace \texttt{<upper>} with a red sweater”) directs edits to the corresponding subject slot. We apply multi-modal classifier-free guidance with a subject-specific scale $s_{\text{entity}}$ inside the mask, keeping the background unchanged. This yields high-fidelity, predictable edits, and also supports reverse (decomposition) tasks by localizing which region to extract or describe. Clean, single-entity frames ensure unambiguous token–region alignment, making control stable and interpretable.

\subsection{Dataset Details}

Our interleaved dataset spans three task families (composition/decomposition, style transfer, and interleaved tutorials) and is constructed via a multi-stage pipeline ensuring high quality.
\vspace{-1.50em}
\paragraph{Dataset Sources.}
Our interleaved dataset integrates multiple real-world tutorial and transformation scenarios from both original in-house creation and carefully selected public resources. 
Sources include: 
(1) high-quality cooking and painting tutorials collected from the internet and manually screened for clarity and completeness; 
(2) public datasets such as RecipeGen, Youcook, MakeAnything, Echo-4o, Viton-HD, Deepfashion, Dresscode, Omnigen and Promptsref website, used strictly within permitted terms; 
\vspace{-1.50em}
\paragraph{Quality Filtering.}
To ensure strong text-image alignment and visual clarity, all sequences are manually or semi-automatically recaptioned using a high-capability multimodal model like GPT-4o and, when needed, re-sampled to preserve temporal and semantic consistency. 
Low-quality or mismatched content is removed during this process so that only well-aligned, contextually coherent, and visually complete examples remain in the final dataset.
\vspace{-1.50em}
\paragraph{Usability.}
The dataset emphasizes realistic, naturally occurring visual content rather than synthetic renderings, ensuring credibility and high transferability to real-world use cases. 
Sequences are paired with step-by-step instructions and corresponding images, providing rich multimodal supervision for interleaved generation tasks. 
We aim to release metadata, filtering scripts, and as many full sequences as licensing permits, enabling reproducibility. 
The resulting corpus offers substantial scale, precise text-image alignment, and high-quality frame progression, making it suitable for both benchmarking and training.

\section{Additional Experiment Results}
\label{app:implementation}

\subsection{Evaluation Metrics Reliability}
\label{sec:EMR}
Our evaluation framework directly targets the key requirements of long-horizon interleaved text–image generation, including : visual continuity (temporal coherence between frames), textual continuity (logical progression of instructions) and cross-modal alignment (semantic match between text and images).  These three dimensions map to the reasoning and rendering challenges in N-to-M generation, providing a balanced view of stepwise planning, visual consistency, and semantic coupling.

\begin{figure*}[!htbp]
    \centering
    \includegraphics[trim=1.00cm 5.55cm 3.95cm 0.10cm, clip, width=0.95\textwidth, page=17]{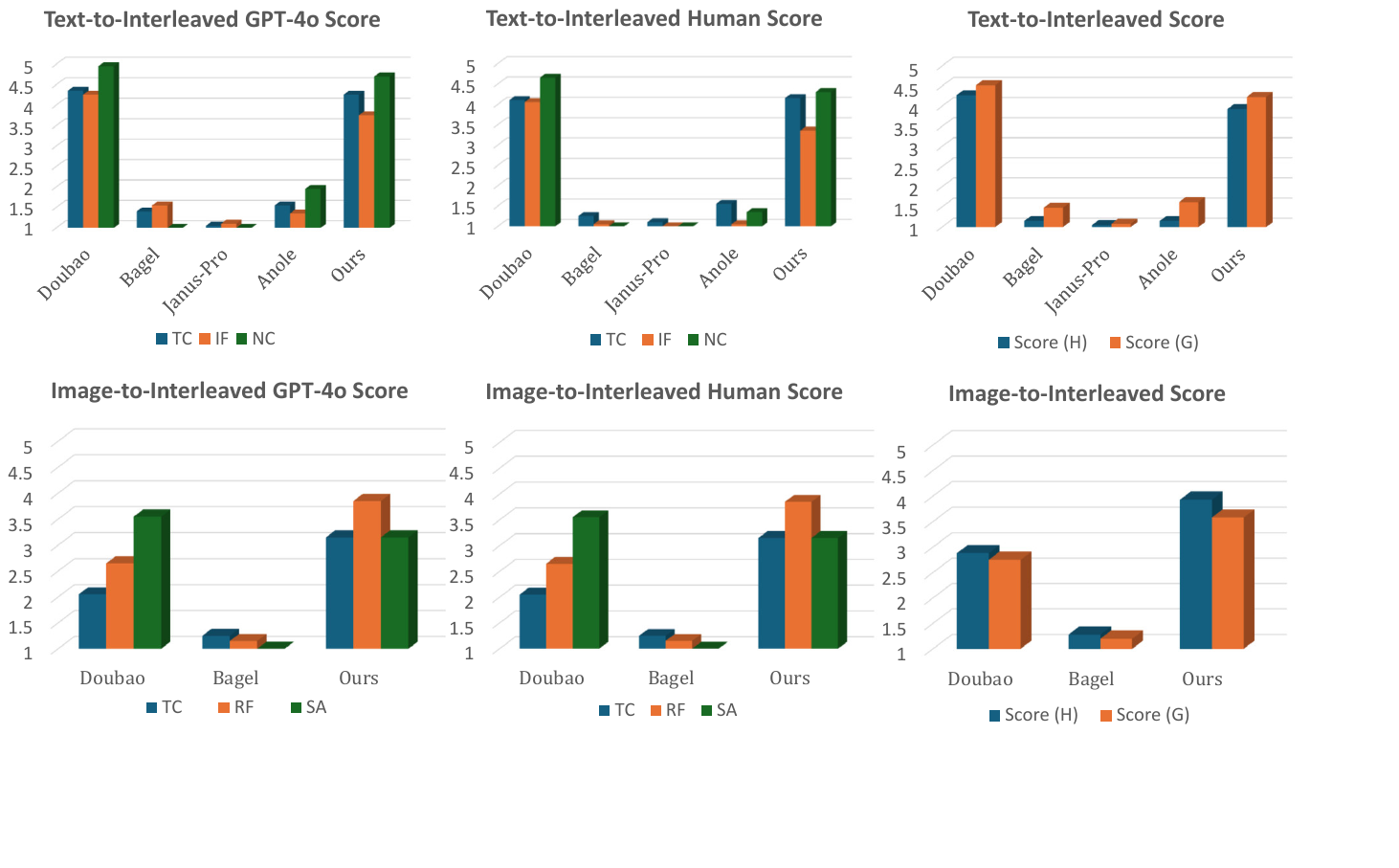}
    \caption{GPT-4o and human scoring details in each metrics and two evaluators comparison. 
    TC. = Temporal Coherence; IF. = Instruction Following; NC. = Narrative Consistency; 
    RF. = Reference Faithfulness; SA. = Semantic Alignment; H = Human; G = GPT-4o.}
    \label{fig:figuresdetails}

    \vspace{0.5em}

    \begin{minipage}{0.48\textwidth}
        \centering
        \scriptsize
        \captionof{table}{Pearson correlation (r) for text-to-interleaved generation criteria. 
        TC. = Temporal Coherence; IF. = Instruction Following; NC. = Narrative Consistency.}
        \label{tab:Pearson_text}
        \renewcommand{\arraystretch}{1.0}
        \begin{tabularx}{\linewidth}{
            >{\centering\arraybackslash}p{1.6cm}  
            | >{\centering\arraybackslash}p{1.6cm} 
            | >{\centering\arraybackslash}p{1.6cm} 
            | >{\centering\arraybackslash}p{1.6cm}
        }
        \toprule
        \textbf{Model} & \textbf{TC.} & \textbf{IF.} & \textbf{NC.} \\
        \midrule
        Doubao & 0.1048 & 0.1325 & 0.3126 \\
        \midrule
        Bagel & 0.2357 & 0.2536 & --   \\
        Janus-Pro & -0.0765 & 0.0526 & --    \\
        Anole & 0.2536 & 0.1683 & 0.1683 \\
        \textbf{Loom (Ours)} & 0.2425 & 0.0605 & 0.2857 \\
        \bottomrule
        \end{tabularx}
    \end{minipage}
    \hfill
    \begin{minipage}{0.48\textwidth}
        \centering
        \scriptsize
        \captionof{table}{Pearson correlation (r) for image-to-interleaved generation criteria. 
        TC. = Temporal Coherence; RF. = Reference Faithfulness; SA. = Semantic Alignment.}
        \label{tab:Pearson_image}
        \renewcommand{\arraystretch}{1.0}
        \begin{tabularx}{\linewidth}{
            >{\centering\arraybackslash}p{1.6cm}  
            | >{\centering\arraybackslash}p{1.6cm}
            | >{\centering\arraybackslash}p{1.6cm} 
            | >{\centering\arraybackslash}p{1.6cm} 
        }
        \toprule
        \textbf{Model} & \textbf{TC.} & \textbf{RF.} & \textbf{SA.} \\
        \midrule
        Doubao & 0.0964 & 0.0989 & 0.183 \\
        \midrule[0.2pt]
        Bagel & 0.1325 & 0.380 & -- \\
        \textbf{Loom (Ours)} & 0.3083 & 0.1765 & 0.0964 \\
        \bottomrule
        \end{tabularx}
    \end{minipage}
\end{figure*}

Given the lack of a public benchmark for extended interleaved generation, we align human evaluation with a strong multimodal LLM evaluator (GPT-4o). Human scoring follows a concise three-question rubric for TC, IF, and NC, parallel to GPT-4o’s evaluation protocol. To verify metric reliability, we compute Pearson correlations between human ratings and GPT-4o scores for text-to-interleaved generation criteria (shown in Fig~\ref{tab:Pearson_text}) and image-to-interleaved generation criteria (shown in Fig~\ref{tab:Pearson_image}). Across all reported dimensions, we find positive correlations and consistent relative rankings, indicating that automatic scores track human preferences well.

A visual comparison between GPT-4o and human scores is shown in Fig~\ref{fig:figuresdetails}. This cross-check ensures that our chosen dimensions are not only intuitive and discriminative, but also statistically aligned between independent evaluators.

In addition, for text-to-interleaved tasks, we report CLIP (ViT-L/14) image–text alignment scores as an auxiliary measure of cross-modal consistency, which correlate positively with human ratings on alignment-related dimensions. Together, these results confirm that our evaluation suite is transparent, task-relevant, and methodologically sound: the three core dimensions cover interleaved generation needs, human and machine assessments exhibit strong Pearson agreement, and CLIP alignment provides an independent signal on image–text coupling.

\subsection{User Study and Evaluation Details}

\begin{figure*}[t]
  \centering
  \includegraphics[trim=1.00cm 9.35cm 8.85cm 0.00cm, clip,width=1.0\textwidth,page=15]{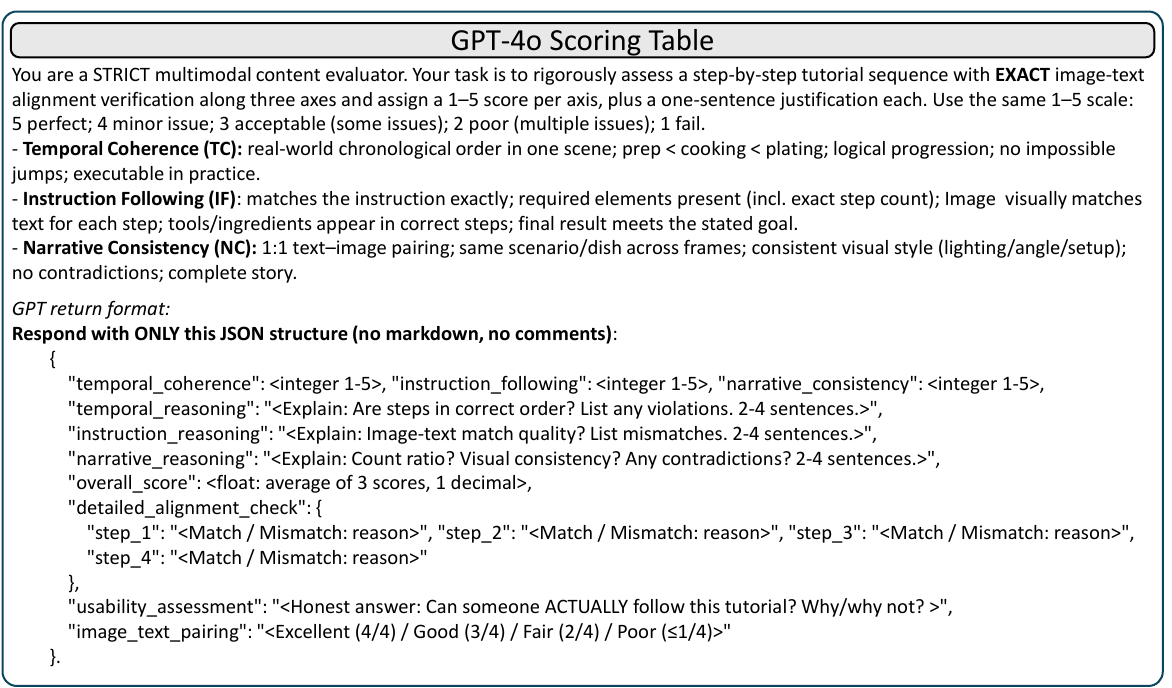}
  \caption{GPT-4o scoring table (automatic evaluation over TC/IF/NC; 1–5, higher is better).}
  \label{fig:gpt-4oscoringtable}
\end{figure*}

\begin{figure*}[t]
  \centering
  \includegraphics[trim=1.00cm 1.05cm 8.85cm 11.75cm, clip,width=1.0\textwidth,page=15]{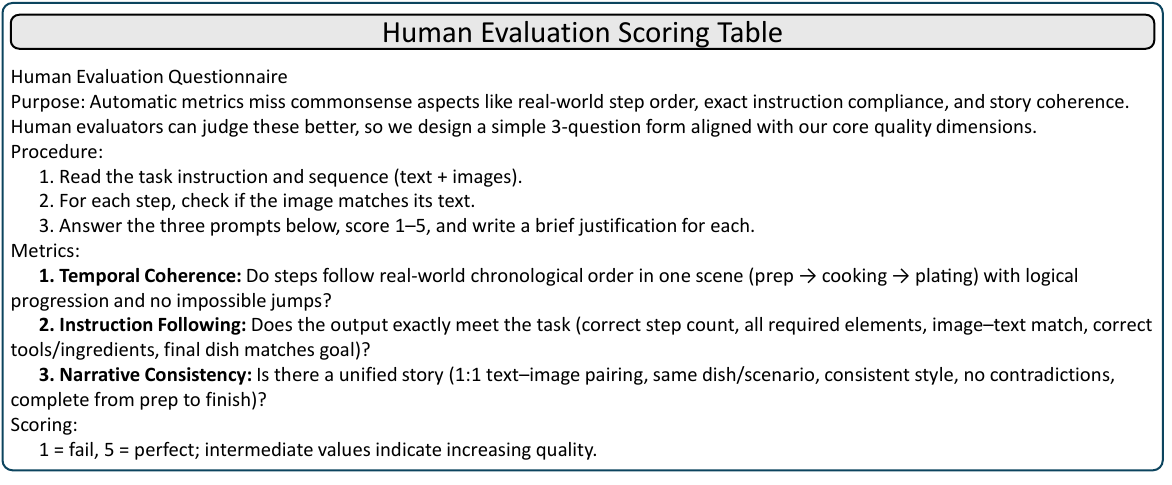}
  \captionsetup{width=1.0\textwidth}
  \caption{Human scoring table (three-question questionnaire for TC/IF/NC; 1–5 with brief rationale).}
  \label{fig:humanscoringtable}
\end{figure*}

GPT-4o scoring details are shown in Fig~\ref{fig:gpt-4oscoringtable}: prompts, 1–5 scoring anchors, and the aggregation rule (mean of TC~/~IF~/~NC). 
Human scoring details are shown in Fig~\ref{fig:humanscoringtable}: questionnaire layout, per-criterion rubric, and reporting fields (rater scores, mean). 
The human study involved 20 participants, each evaluating 20 assigned test cases; for each assigned case, a participant scored all model outputs for that case, without evaluating every case in the dataset.

\subsection{More Generation Results}

\begin{figure*}[t]
  \centering
  \includegraphics[trim=0.05cm 5.15cm 1.00cm 0.25cm, clip, width=1.0\textwidth, page=4]{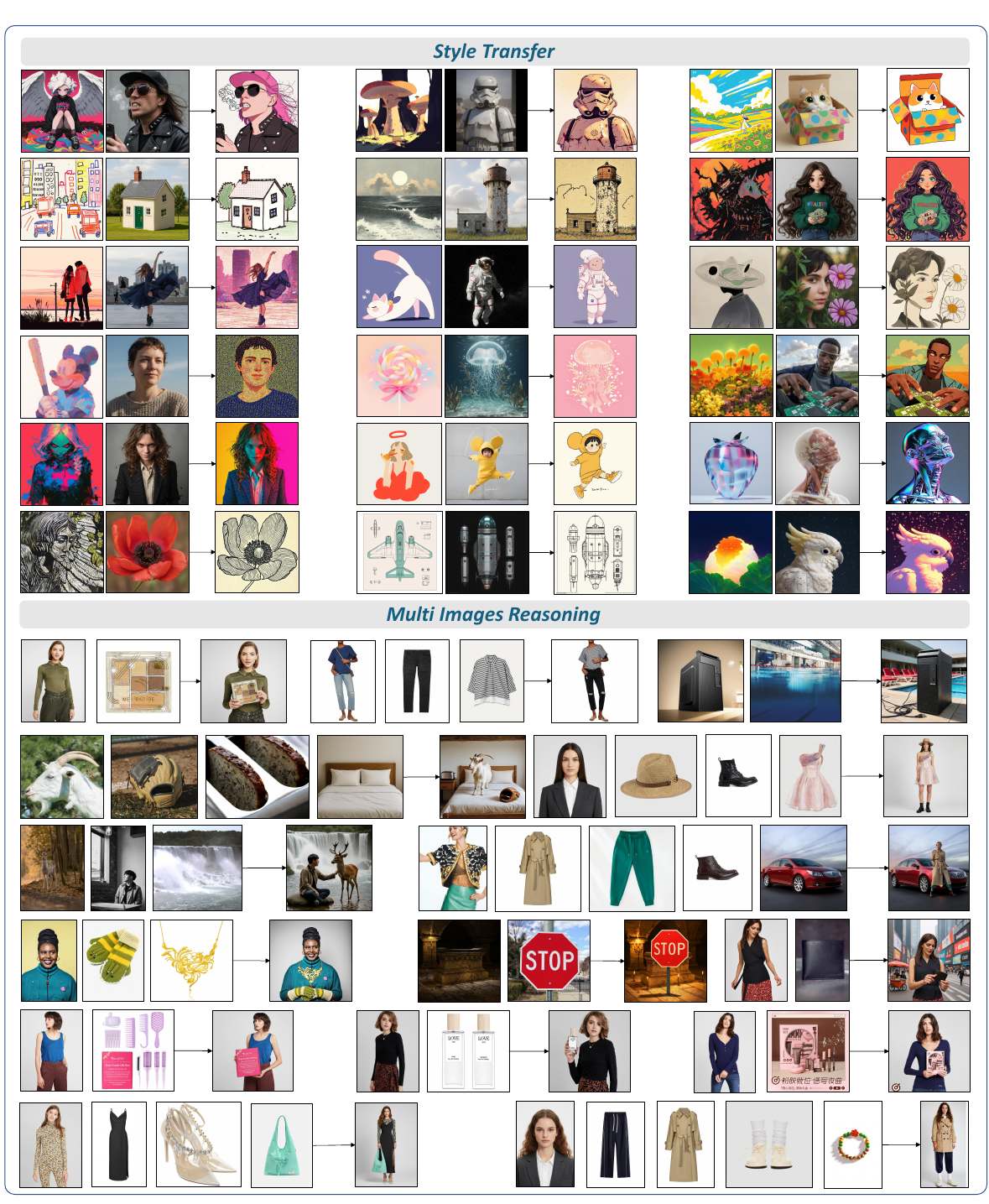}
  \caption{More generation results of Loom in interleaved input tasks, which include multi images reasoning and style transfer.}
  \label{fig:formoreinputresults}
\end{figure*}

\begin{figure*}[t]
  \centering
  \includegraphics[trim=0.05cm 5.15cm 1.00cm 0.25cm, clip, width=1.0\textwidth, page=5]{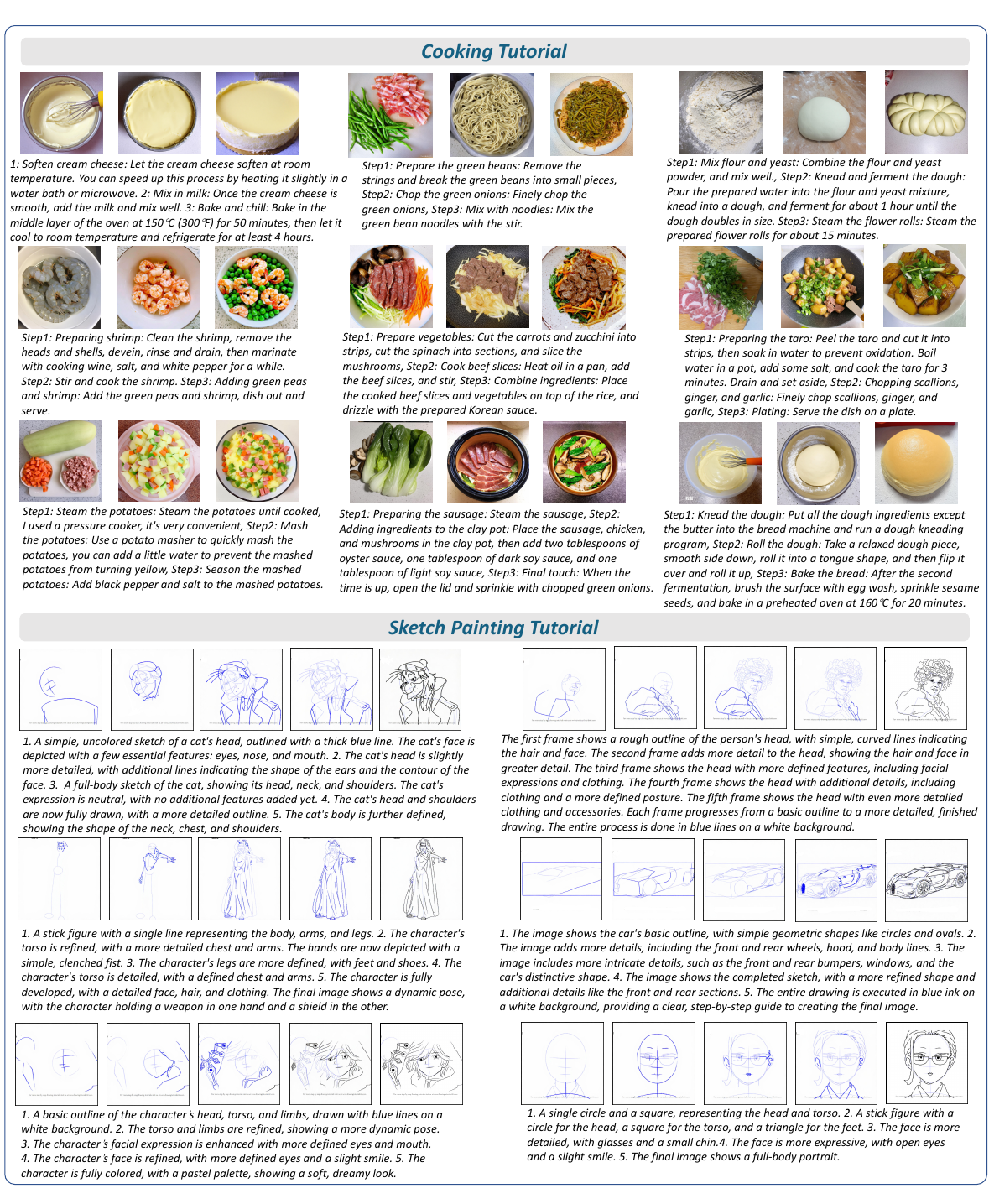}
  \caption{More generation results of Loom in interleaved output tasks, which include cooking tutorial and sketch painting tutorial.}
  \label{fig:formoreoutputresults}
\end{figure*}

More interleaved input and output results shown in Fig~\ref{fig:formoreinputresults} and Fig ~\ref{fig:formoreoutputresults}.

\section{Limitation and Future Work}

While Loom establishes a strong baseline for open-source interleaved generation, a performance gap remains compared to proprietary systems. Loom currently prioritizes temporal consistency and instruction adherence, trading off single-frame visual fidelity against specialized image synthesizers. Furthermore, Loom trails closed-source counterparts in complex reasoning tasks. Future directions include scaling the reasoning backbone, enhancing high-resolution synthesis, and expanding tutorial datasets to close the gap with proprietary frontiers.

\end{document}